\documentclass[10pt,twocolumn,letterpaper]{article}
\usepackage{comment}
\usepackage{placeins}
\usepackage{bm}
\usepackage[pagenumbers]{cvpr}      
\usepackage[linesnumbered,ruled]{algorithm2e}
\usepackage{multicol}
\usepackage{url}
\usepackage{comment}
\usepackage{array} 
\usepackage{arydshln} 
\usepackage{pifont}
\usepackage{amssymb}
\usepackage[table]{xcolor}

\usepackage{siunitx} 
\usepackage{booktabs} 

\usepackage{placeins}

\definecolor{cvprblue}{rgb}{0.21,0.49,0.74}
\usepackage[pagebackref,breaklinks,colorlinks,allcolors=cvprblue]{hyperref}

\def\paperID{*****} 
\def\confName{CVPR}
\def\confYear{2026}

\title{Image-based Joint-level Detection for Inflammation in Rheumatoid Arthritis from Small and Imbalanced Data}

\author{Shun Kato\\
Keio University\\
{\tt\small kato\_shun1329@keio.jp}
\and
Yasushi Kondo\\
Keio University\\
{\tt\small yasutonko@keio.jp}
\and
Shuntaro Saito\\
Keio University\\
{\tt\small shun081359@gmail.com}
\and
Yoshimitsu Aoki\\
Keio University\\
{\tt\small aoki@elec.keio.ac.jp}
\and
Mariko Isogawa\\
Keio University\\
{\tt\small isogawa@ics.keio.ac.jp}
}
\usepackage{times}
\usepackage{epsfig}
\usepackage{graphicx}
\usepackage{amsmath}
\usepackage{amssymb}

\usepackage{comment}

\begin{document}
\maketitle
\begin{abstract}
Rheumatoid arthritis (RA) is an autoimmune disease characterized by systemic joint inflammation. 
Early diagnosis and tight follow-up are essential to the management of RA, as ongoing inflammation can cause irreversible joint damage.
The detection of arthritis is important for diagnosis and assessment of disease activity; however, it often takes a long time for patients to receive appropriate specialist care. Therefore, there is a strong need to develop systems that can detect joint inflammation easily using RGB images captured at home. 
Consequently, we tackle the task of RA inflammation detection from RGB hand images.
This task is highly challenging due to general issues in medical imaging, such as the scarcity of positive samples, data imbalance, and the inherent difficulty of the task itself.
However, to the best of our knowledge, no existing work has explicitly addressed these challenges in RGB-based RA inflammation detection.
This paper quantitatively demonstrates the difficulty of visually detecting inflammation by constructing a dedicated dataset, and we propose a inflammation detection framework with global local encoder that combines self-supervised pretraining on large-scale healthy hand images with imbalance-aware training to detect RA-related joint inflammation from RGB hand images.
Our experiments demonstrated that the proposed approach improves F1-score by 0.2 points and Gmean by 0.25 points compared with the baseline model.
\end{abstract}

\section{Introduction}
\label{sec:intro}
Rheumatoid arthritis (RA) is a common, while refractory autoimmune disease that characterized by synovial inflammation that results in systemic joint damage. Such joint damage can cause severe bone destruction and prevent patients from moving their hands freely.

Recent advances in the development of new therapeutics, including biologics and Janus kinase inhibitors have made it possible to aim for remission as a primary goal of management of RA~\cite{kondo-paradigm}. However, missing timely access to appropriate medical care can delay the diagnosis of RA or lead to inadequate control of disease activity. 

Therefore, detecting active synovial inflammation is critically important in RA management. It is an early manifestation of the disease.
However, this requires assessment by rheumatologists and the utilisation of specialised equipment including joint ultrasound, if the inflammation is subtle~\cite{diagnosis-ra}. Whereas, the number of rheumatologists is limited and unevenly distributed across regions, RA patients may not receive appropriate assessment of joint inflammation in their family medicine clinic or have to make a long journey to find out specialists. Therefore, there is a great need for development in technologies that can support remote RA diagnosis and management of disease activities though the detecting active joint inflammations.

However, traditional methods for diagnosing RA are known to be poorly suited for telemedicine. Accurate diagnosis has typically relied on X-ray~\cite{X-ray-kantsu, pitagorasu-X-ray, hand-xray-transfer} or MRI~\cite{MRI-1, MRI-ultrasound} imaging, which require specialized equipment and expertise. As a result, these methods are generally limited to patients who can visit medical facilities. Therefore, for RA detection in telemedicine, it is important to use modalities that do not require specialized equipment or expertise (i.e., RGB images) that can be easily captured by the general patients.

Therefore, this paper addresses the detection of RA inflammation in individual finger joints using only RGB images captured by conventional cameras, without the need for specialists or specialized equipment. With this method, a patient experiencing joint pain can take a photo of their fingers to detect the exists of the disease. The patient can use the results to decide whether to visit a hospital or as a reference for receiving remote medical care.

However, following three technical challenges exist in RA inflammation detection from RGB hand images.
\begin{itemize}
    \item{
    Firstly, as with many other medical imaging tasks, it is difficult to collect hand and finger image samples from affected patients at scale.
    Consequently, we need to learn discriminative characteristics of diseased hands from small dataset.
    }
    \item{
    One practical way to address the small dataset is to leverage publicly available hand and finger image datasets; however, almost all images are healthy. Therefore, even if such images can be used for training, it remains necessary to handle the resulting class imbalance.
    }
    \item{
    The third factor is the lack of highly reliable datasets. It is known that rheumatologists' visual inspection does not always agree with the actual presence or absence of inflammation~\cite{correlation}; therefore, ground-truth labels for inflammation should be assigned based on ultrasound findings rather than visual inspection. However, to the best of our knowledge, datasets annotated in this manner are currently unavailable.
    }
\end{itemize}
Therefore, to address these challenges, we propose a robust deep learning-based approach for joint-level RA inflammation detection. The main contributions of this work are as follows:
\noindent
(1) We propose a novel deep learning-based framework that providing a scalable solution for remote RA monitoring. (2) We leverage a pre-training strategy using a large-scale healthy hand dataset to learn informative and robust hand feature representations, effectively overcoming the inherent challenges of training with a limited number of RA-positive samples. (3) We introduce an imbalance-aware optimization strategy by incorporating Focal Loss to mitigate the severe class imbalance typical of medical datasets, ensuring high sensitivity even in joints with low inflammation prevalence. (4) We construct a high-reliability RGB hand image dataset where inflammation labels are assigned based on ultrasound findings rather than subjective visual inspection. To the best of our knowledge, this is the first study to provide an objective dataset that bridges the gap between external appearance and internal synovial inflammation.

\section{Related Works}
\label{sec:related}
\subsection{Detection for joint RA inflammation}
Current standard way of detection for finger joint inflammation in RA relies primarily on physical examination by a rheumatologist. Furthermore, as mentioned in the introduction, detecting subtle arthritis often requires MRI~\cite{MRI-1} or joint ultrasound~\cite{MRI-ultrasound}, necessitating a visit to a specialized medical facility.

Regarding the research that related to RA telemedicine using imaging technology, the detection of bone erosions using conventional X-rays has been proposed. However, these approaches primarily evaluate structural changes resulting from past inflammation, rather than the active joint inflammation targeted in our study. In addition, recent studies have explored the utilizing of thermography as a tool for detecting of joint inflammation, referencing MRI synovitis. However, the main problem is the general applicability of thermography devices as a modality of telemedicine. 

Compared to these other modalities, RGB image-based methods have the potential to contribute to telemedicine. However, unlike X-ray or MRI, RGB images can only capture the external appearance of the body, making diagnosis based solely on these images a highly challenging task.
Notably, Phatak et al.~\cite{pathak} pioneered this field by performing joint-level detection of RA inflammation using InceptionResNetv2~\cite{inceptionresnetv2}, thereby validating the potential of CNN-based approaches for this task.
However, visual signs may not always perfectly reflect active inflammation~\cite{correlation}, establishing a more robust ground truth would be beneficial.

\subsection{Machine Learning Methods for Small/Imbalanced Datasets}
In image classification tasks, class imbalance of dataset is a very common and significant issue.
Class Imbalance causes biased classification and overfitting, and leads to a degraderation of the model performance.
Major approaches to address class imbalance are class resampling~\cite{ADASYN,SMOTE, BSMOTE, under-vs-over} and cost-sensitive training~\cite{cost-sesitive,syuuseiCE, MFE, imbalanced-daiic, focalloss}.

Class resampling addresses class imbalance by adjusting the dataset through methods such as undersampling, which reduces majority class samples, or oversampling, which increases minority class samples. In contrast, cost-sensitive training tackles class imbalance by assigning different weights to loss functions, making the model more responsive to minority classes. However, these approaches are limited in their effectiveness when the original dataset is extremely small, as they can't address the issue of insufficient data volume.

In medical image classification, the number of positive samples in a dataset tends to be extremely limited because only a small number of individuals recognize their symptoms at an early stage and seek medical attention. As a result, many datasets become small in size and exhibit class imbalance. However, to the best of our knowledge, existing studies on the detection of RA inflammation using machine learning-based image classification have not specifically addressed the issue of imbalanced datasets.

\section{Proposed Method}
\begin{figure*}[t]
  \centering
  \includegraphics[width=1.0\textwidth]{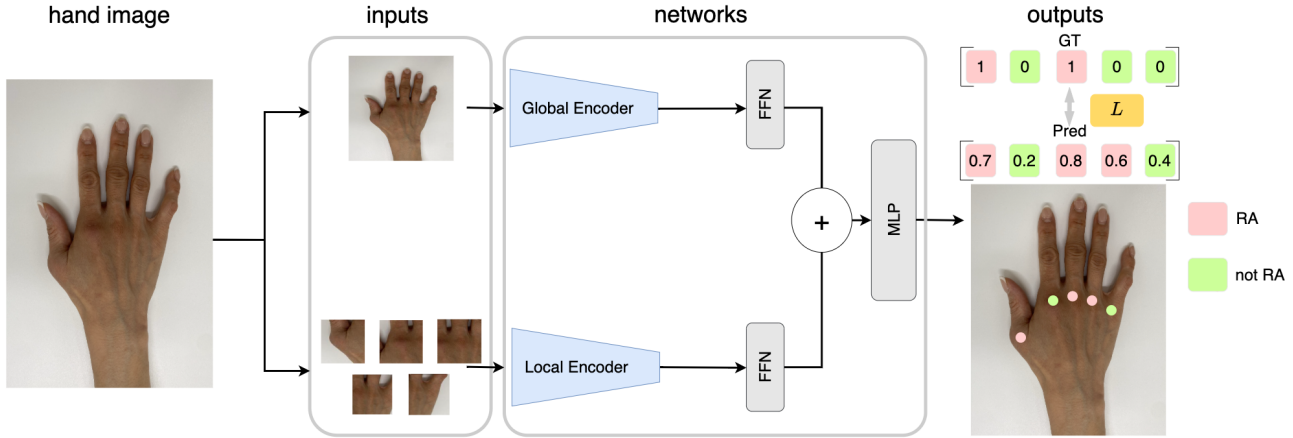} 
  \caption{Overview of the proposed framework. Joint regions are cropped from the input hand image and fed into the Local Encoder, while the whole hand image is input to the Global Encoder. The model estimates the presence of inflammation at the joint level by integrating global features from the whole hand with local features from the joints.}
  \label{proposed-framework}
\end{figure*}
\subsection{Overview of the Proposed Framework}
To robustly estimate the presence of Rheumatoid Arthritis (RA) joint inflammation from RGB images, we propose a novel framework that integrates both global and local visual features. Figure~\ref{proposed-framework} illustrates the overall pipeline of our method. 

Given a single RGB image of the hand $X$, our goal is to output a binary label $\bm{\mathit{y}} = [y_{1}, y_{2}, \ldots, y_{N}]$ indicating the presence or absence of joint inflammation for each hand joint, where $N$ denotes the number of joints. To achieve this, our framework consists of two main phases:\\
\noindent
\textbf{Feature Extraction via Global-Local Encoders:} We utilize an image recognition model equipped with Global and Local Encoders to capture both overall hand deformities and specific joint-level features from cropped joint images $\bm{\mathit{X_{local}}} = [x_{1}, x_{2}, \ldots, x_{N}]$.\\
\noindent
\textbf{Two-Stage Training Strategy:} To overcome the scarcity of positive data and extreme class imbalance, we conduct a two-stage training approach. This involves self-supervised pre-training on a large-scale hand dataset, followed by fine-tuning using Focal Loss~\cite{focalloss}.

Detailed descriptions of the network architecture and the training strategy are provided in the following subsections.

\subsection{Network Architecture}
To effectively capture the subtle visual cues of inflammation, our network consists of two parallel encoders. Specifically, we build two types of encoders based on ResNet-18~\cite{resnet18}:
\begin{itemize}
    \item \textbf{Global Encoder:} Processes the full hand image $X$ to capture overall features, including global deformities and structural contexts.
    \item \textbf{Local Encoder:} Processes the cropped images $\bm{\mathit{X_{local}}}$ of each joint region. For cropping, a hand joint position estimation framework is utilized to extract patch images using the detected joint positions as center coordinates.
\end{itemize}
The outputs from both encoders are then integrated and fed into the MLP header to classify the inflammatory status for each joint.

\subsection{Two-Stage Training Strategy}

\subsubsection{Self-Supervised Pre-training}
To address the lack of annotated RA images and to ensure robust feature extraction, both the Global and Local Encoders are pre-trained before tackling the target task.
We trained ResNet-18 backbones using DINO~\cite{dino}
 on the large-scale 11k-hands dataset~\cite{11k-hands}. This stage allows the encoders to learn rich, generalized representations of hand anatomy without requiring labeled disease data.

\subsubsection{Fine-tuning with Focal Loss}
In the second stage, we fine-tune the network on our specific RA dataset. To prevent the pre-trained representations from degrading and to mitigate the risk of overfitting, we freeze the weights of both the Global and Local Encoders. Only the attached FFN and the integrating MLP header are set as trainable.

Furthermore, to deal with the severe data imbalance, we train the network using Focal Loss~\cite{focalloss} instead of standard Binary Cross Entropy (BCE). The objective function is defined as:
\begin{align*}
\mathcal{L}(\theta)
&= - \frac{1}{M}\sum_{i=1}^{M} \Big[
y_i (1-\hat{y}_i)^{\gamma}\log(\hat{y}_i) \nonumber\\
&\qquad\qquad + (1-y_i)\hat{y}_i^{\gamma}\log(1-\hat{y}_i)
\Big]
\end{align*}
where $\theta$ denotes all trainable parameters in the header, $M$ is the number of samples, and $\hat{y}_{i}$ and $y_{i}$ represent the predicted probability and the ground truth label, respectively. The focusing parameter $\gamma \geq 0$ adjusts the rate at which easy examples are down-weighted. 
\section{Experiments}
\label{sec:exps}
\subsection{Dataset}
\FloatBarrier

\begin{table}[t]
  \centering
  \caption{Comparison between our dataset and existing work ~\cite{b35}}
  \label{comparison_with_pathak}
  \setlength{\tabcolsep}{3pt}
  \renewcommand{\arraystretch}{1.05}
  \begin{tabular}{lccccc}
    \toprule
    dataset & year & patients & hand & joint & ultrasound \\
    \midrule
    Phatak et al.~\cite{b35} & 2024 & 200 & 400 & 5600 & \ding{55} \\
    Ours & 2025 & 68 & 222 & 2392 & \checkmark \\
    \bottomrule
  \end{tabular}
\end{table}

\renewcommand{\labelitemi}{$\bullet$}
\noindent{\bf{Rheumatoid Arthritis Dataset}}: 
To evaluate our proposed framework and baseline models, we collected an original RGB hand image dataset containing inflammation caused by Rheumatoid Arthritis.
Table~\ref{comparison_with_pathak} shows a comparison with the dataset of exsisting work. Although our dataset is smaller in scale compared to Phatak et al.~\cite{b35}, which relies on visual cues, our annotations were derived from ultrasound diagnostic findings, ensuring high reliability.
\begin{figure*}[t]
  \centering
  \begin{minipage}[b]{0.49\textwidth}
    \centering
    \includegraphics[width=\linewidth]{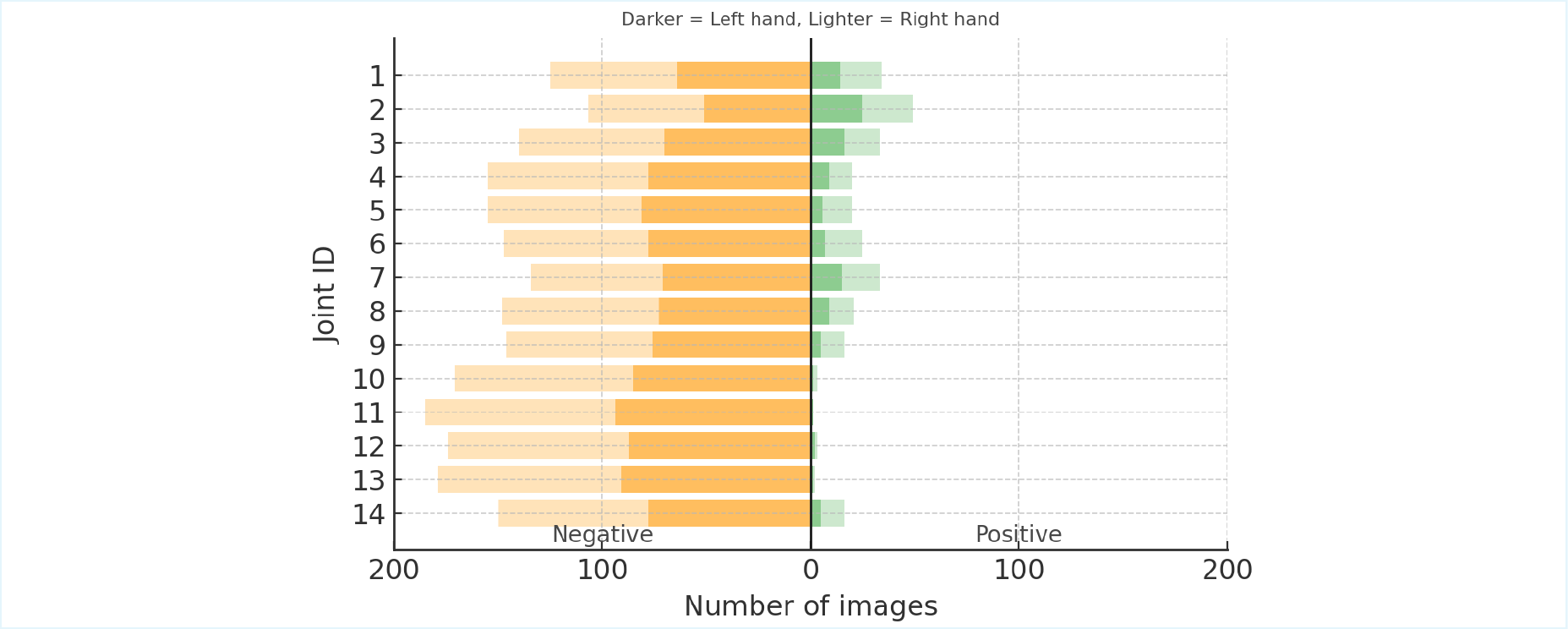}
    \caption{Number of positive cases at each finger joint. A 14-joint hand skeleton based on MCP/PIP/DIP joints.}
    \label{positive_rate_of_dataset}
  \end{minipage}
  \hfill
  \begin{minipage}[b]{0.49\textwidth}
    \centering
    \includegraphics[width=\linewidth]{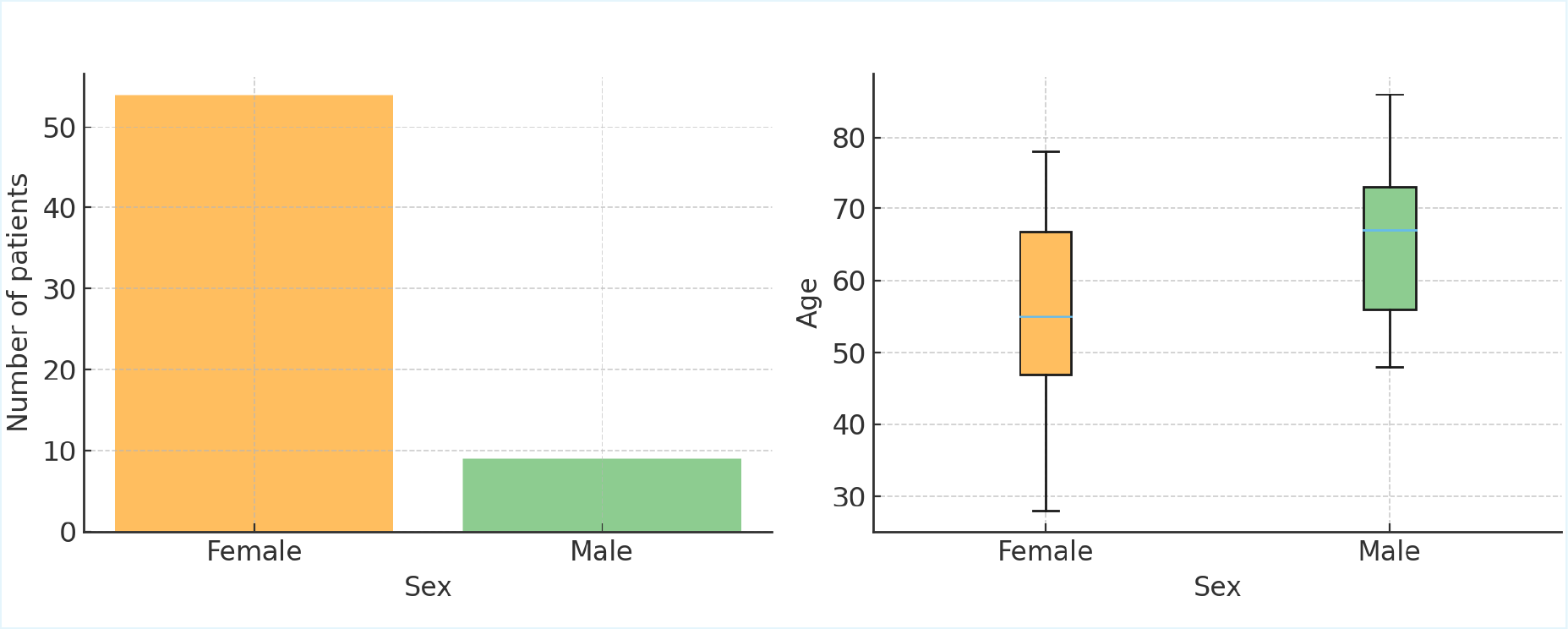}
    \caption{Distributions of sex and age in the dataset. Blue line represents the mean value.}
    \label{dataset_statistics}
  \end{minipage}
\end{figure*}
\noindent
{\bf{Ethics}}
This study was approved by Keio University School of Medicine Ethics Committee with approval number of 20200262, 20211061 and conformed to the 2008 Declaration of Helsinki and the 2008 Ethical Guidelines for Clinical Research by the Japanese Ministry of Health, Labour, and Welfare. All patients provided written informed consent.

\noindent
{\bf{Patients and Statistics}}:
Patients with inflammatory arthritis of the hand and finger joints were enrolled in the outpatient rheumatology clinic at Keio University, Japan. The patients were followed for 12 weeks and hand images were captured at week 0 and at week 12.
A rheumatologist instructed each patient to fully extend the hand and fingers, then photographed both hands from the dorsal aspect. All photographs were taken using an iPhone 11(Apple Inc., Cupertino , CA, USA).
Furthermore, to focus on early stage inflammation, we excluded cases that had progressed to ankylosis or joint destruction based on the rheumatologist's assessment.
Consequently, we captured 222 hand images from 68 patients.

The distribution of patient sex and age is shown in Figure~\ref{dataset_statistics}. Interestingly, a larger proportion of the enrolled patients were of female sex, which is consistent with higher prevalence of rheumatoid arthritis in females.

~Figure~\ref{positive_rate_of_dataset} presents the number of positive and negative samples for each joint in the dataset. Even in patients diagnosed with RA, inflammation typically does not manifest in all joints simultaneously but is localized to specific joints. Consequently, at the joint level, the number of positive samples is significantly lower than negative ones, resulting in a high class imbalance. This is particularly evident in DIP joints, where some joint points contain zero positive samples.

\noindent
{\bf{Annotation}}
We annotated the presence or absence of synovial inflammation based on ultrasound findings. Ultrasound examinations were performed by the rheumatologist and, cases meeting gray scale(GS) $>$ 1 and power doppler(PD) $>$ 0, were labeled as inflamed, while all other cases were labeled as non-inflamed.
Consequently, each data entry consists of a unique image paired with a binary label for inflammation derived from the ultrasound assessment.

\subsection{Evaluation Metrics}
%
Under conditions of class imbalance, classification accuracy often overestimates the performance of models, as it favors majority class predictions. Therefore, we employed Recall, Precision, F1-score, and Geometric Mean (Gmean) as metrics. Recall evaluates the model's ability to identify all relevant positive samples, while Precision measures the correctness of the positive predictions. The F1-score, defined as the harmonic mean of Precision and Recall, and Gmean measures the geometric mean of sensitivity and specificity, ensuring that the model maintains a balanced accuracy across all classes.

\subsection{Implementation Details}
{\bf{Dataset Preprocessing}}:
To feed into the network, we applied background removal. In particular, we conducted segmentation of the hand using SAM 2~\cite{SAM2} and padded background to zero.
In addition, the captured images were processed with MediaPipe~\cite{mediapipe} to detect 14 joint landmarks. Patch images were extracted using the detected joint landmarks as center coordinates, ensuring that each joint is adequately captured. 
The whole hand image was resized to $224 \times 224$ pixels, while cropped images $\bm{\mathit{X_{local}}}$ initially cropped to 64 pixels square and then resized to $224 \times 224$ pixels. Then we excluded Distal Interphalangeal (DIP) joints from our dataset, as they are known to have an extremely low prevalence of inflammation.
To ensure a robust evaluation, we conducted a 5-fold cross-validation. In this setup, the dataset was divided into training and testing sets with an 8:2 ratio for each fold. We report the average performance across all five folds.

\noindent
{\bf{Training Setup}}:
We used ResNet-18 for both Global and Local Encoder. During the pre-training, both the Global and Local Encoders were trained with the learning rate of $1.5e-4$, while the fine-tuning stage employed $1e-4$ and $2.0$ for learning rate and gamma parameter, respectively. All baseline models were initialized with ImageNet~\cite{imagenet} pretrained weights and fine-tuned on our dataset.
For the linear probing experiment with the Vision Transformer~\cite{vit}, we trained only the classifier head.
We employed AdamW~\cite{AdamW} for pretraining and Adam~\cite{Adam} for fine-tuning as the optimizer. We trained our model on a RTX Ada Generation with 48GB of GPU memory.

\subsection{Experiments and Results}

\begin{table}[t]
\centering
\caption{Comparison with baseline methods.}
\label{tab:results}
\renewcommand{\arraystretch}{1.2}
\small
\setlength{\tabcolsep}{3pt} 
\begin{tabular}{@{\extracolsep{\fill}}lcccc@{\extracolsep{\fill}}}
\toprule
\textbf{Model} & \textbf{Recall} & \textbf{Precision} & \textbf{F1} & \textbf{Gmean} \\
\midrule
Rheumatologists (Avg.) & 0.583 & 0.214 & 0.305 & 0.623 \\
\midrule
ResNet-18~\cite{resnet18} & 0.170 & 0.358 & 0.230 & 0.356 \\
ResNet-50~\cite{resnet18} & 0.155 & 0.301 & 0.205 & 0.343 \\
Phatak et al.~\cite{b35} & 0.129 & \textbf{0.601} & 0.212 & 0.351 \\
ViT-B/16 (FFT)~\cite{vit} & 0.000 & 0.000 & 0.000 & 0.000 \\
ViT-B/16 (Linear Prob.) & 0.000 & 0.000 & 0.000 & 0.000 \\
\midrule
\textbf{Ours} & \textbf{0.478} & 0.374 & \textbf{0.420} & \textbf{0.605} \\
\bottomrule
\end{tabular}
\end{table}

\begin{table}[t]
\centering
\caption{Ablation study on key components. Each component contributes to the overall performance.}
\label{tab:ablation}
\renewcommand{\arraystretch}{1.2}
\small
\setlength{\tabcolsep}{3pt}
\begin{tabular}{@{\extracolsep{\fill}}lcccc@{\extracolsep{\fill}}}
\toprule
\textbf{Variant} & \textbf{Recall} & \textbf{Precision} & \textbf{F1} & \textbf{Gmean} \\
\midrule
\textbf{Ours} & \textbf{0.478} & 0.374 & \textbf{0.420} & \textbf{0.605} \\
\midrule
w/o DINO pre-training & 0.383 & 0.326 & 0.352 & 0.580 \\
w/o Focal Loss & 0.115 & \textbf{0.465} & 0.158 & 0.286 \\
w/o Global/Local Encoder & 0.454 & 0.370 & 0.408 & 0.517 \\
\bottomrule
\end{tabular}
\end{table}
\begin{table}[t]
\centering
\caption{Detailed performance analysis of individual rheumatologists.}
\label{tab:human_experts}
\renewcommand{\arraystretch}{1.2}
\small
\setlength{\tabcolsep}{4pt}
\begin{tabular}{@{\extracolsep{\fill}}lcccc@{\extracolsep{\fill}}}
\toprule
\textbf{Observer} & \textbf{Recall} & \textbf{Precision} & \textbf{F1} & \textbf{Gmean} \\
\midrule
Rheumatologist A & 0.641 & 0.257 & 0.364 & 0.693 \\
Rheumatologist B & 0.558 & 0.184 & 0.250 & 0.587 \\
Rheumatologist C & 0.552 & 0.201 & 0.303 & 0.591 \\
\midrule
Average & 0.583 & 0.214 & 0.305 & 0.623 \\
\bottomrule
\end{tabular}
\end{table}
We conducted two different experiments to investigate our method's efficacy: (1) a comparison against baseline methods, (2) an ablation study to show the importance of the pre-training, Focal Loss and Global Encoder. (3) a comparison between human experts.

\noindent
{\bf{Comparison against baseline methods.}}
Table~\ref{tab:results} shows the quantitative results. Our method outperformed other baselines across almost all metrics, with the exception of Precision compared to InceptionResNetv2. However, as previously discussed, the risk associated with false negatives outweighs that of false positives in clinical diagnosis. Therefore, our framework is considered superior to InceptionResNetv2, which suffers from low Recall.

Throughout the experiments, ResNet-18 consistently outperformed ResNet-50. In contrast, ViT failed to converge properly under both Full Fine-Tuning and Linear Probing settings, resulting in a trivial solution where all instances were predicted as negative.
These results denote that large scale models with high parameter counts are prone to overfitting or model collapse when trained on small scale and imbalanced datasets. In particular, the failure of ViT in Linear Probing highlights the significant domain gap between ImageNet~\cite{imagenet} pre-trained features and our dataset, making effective adaptation difficult.

\noindent
{\bf{Ablative Analysis. }}
This ablation test investigated the effect of (a) pre-training on large-scale hand datasets through self-supervised learning, (b) fine-tuning with the Focal Loss, and (c) global local encoder architecture which are the part of our technical contributions. As shown in the Table~\ref{tab:ablation}, our model reported the best score compared with both our model without the pre-training (denoted as w/o DINO pre-training) and the Focal Loss (denoted as w/o Focal Loss), and the Global/Local Encoder (denoted as w/o Global/Local Encoder) indicating that both essences contributed to improving the estimation accuracy.

\noindent
{\bf{Comparison with Human Experts.}}
~Table~\ref{tab:human_experts} shows a performance comparison of three rheumatologists using our dataset. In this experiment, the rheumatologists were presented solely with joint images and asked to determine the presence of RA inflammation.
Consequently, their Precision consistently remained around 0.2, and F1 scores were limited to about 0.3. This suggests that aligning visual judgment alone with ultrasound based labels is extremely challenging. Given that routine clinical practice involves palpation alongside visual inspection, these results are considered reasonable for a task relying solely on image data.
On the other hand, the rheumatologists demonstrated a relatively high Recall of approximately 0.6. In a clinical setting, the risk of false negatives outweighs that of false positives. Therefore, this can be interpreted as a safety bias, where rheumatologists tend to classify ambiguous cases as positive.

\section{Conclusion and Future Work}
In this paper, we propose a novel framework for RGB based detection of RA related joint inflammation that enables robust estimation even from small and imbalanced datasets.
Experimental results demonstrated that our method quantitatively outperforms other baseline methods.
Our approach not only enables robust learning from imbalanced, limited data but also offers easy adaptability to other diseases affecting finger joints.

As a limitation, this paper assumes that symptoms of rheumatoid arthritis appear as visible changes in RGB images. However, in extremely early cases, changes are hardly visible to the naked eye. Addressing this issue is part of our future work.

\end{document}